\documentclass{article}

\usepackage{arxiv}

\usepackage[utf8]{inputenc} 
\usepackage[T1]{fontenc}    
\usepackage{hyperref}       
\usepackage{url}            
\usepackage{booktabs}       
\usepackage{amsfonts}       
\usepackage{nicefrac}       
\usepackage{microtype}      
\usepackage{graphicx}
\usepackage{natbib}
\usepackage{doi}

\usepackage{multirow}
\usepackage{makecell}
\usepackage{amsmath}
\usepackage{amssymb}
\usepackage{subcaption}
\usepackage{adjustbox}
\usepackage{siunitx}
\usepackage{float}
\usepackage{xcolor}
\usepackage{listings}
\title{Interpretable EEG Microstate Discovery via Variational Deep Embedding: A Systematic Architecture Search with Multi-Quadrant Evaluation}

\author{
    \href{https://orcid.org/0009-0007-8428-7019}{\includegraphics[scale=0.06]{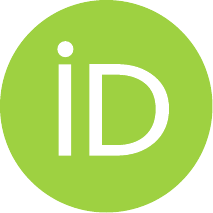}\hspace{1mm}Saheed Faremi}\thanks{Corresponding author. These authors contributed equally.} \\
    School of Computer Science and Information Technology\\
    University College Cork\\
    Western Road, Cork, T12 XF62, Ireland \\
    Artificial Intelligence and Cognitive Load Research Lab,\\
    University College Cork, Ireland \\
    \texttt{126100912@umail.ucc.ie} \\
    \And
    \href{https://orcid.org/0000-0003-3702-4826}{\includegraphics[scale=0.06]{orcid.pdf}\hspace{1mm}Andrea Visentin}\thanks{These authors contributed equally.} \\
    School of Computer Science and Information Technology\\
    University College Cork\\
    Western Road, Cork, T12 XF62, Ireland \\
    Insight RI Research Centre for Data Analytics,\\
    University College Cork, Ireland \\
    \texttt{andrea.visentin@ucc.ie} \\
    \And
    \href{https://orcid.org/0000-0002-2718-5426}{\includegraphics[scale=0.06]{orcid.pdf}\hspace{1mm}Luca Longo}\footnotemark[2] \\
    School of Computer Science and Information Technology\\
    University College Cork\\
    Western Road, Cork, T12 XF62, Ireland \\
    Artificial Intelligence and Cognitive Load Research Lab,\\
    University College Cork, Ireland \\
    \texttt{luca.longo@ucc.ie} \\
}

\renewcommand{\shorttitle}{Interpretable EEG Microstate Discovery via Conv-VaDE}

\hypersetup{
pdftitle={Interpretable EEG Microstate Discovery via Variational Deep Embedding},
pdfsubject={cs.LG, q-bio.NC, eess.SP},
pdfauthor={Saheed Faremi, Andrea Visentin, Luca Longo},
pdfkeywords={EEG microstates, Variational autoencoder, Gaussian mixture model, Polarity invariance, Hyperparameter sweep, Clustering validation, Mechanistic Interpretability},
}

\begin{document}
\maketitle

\begin{abstract}
EEG microstate analysis segments continuous brain electrical activity into brief, quasi-stable topographic configurations that reflect discrete functional brain states. Conventional approaches such as Modified K-Means operate directly in electrode space with hard assignment, offering no learned latent representation, no generative decoder, and no mechanism to decode latent configurations into verifiable scalp topographies, limiting both model transparency and interpretability. To address this, we present a Convolutional Variational Deep Embedding (Conv-VaDE) model that jointly learns topographic reconstruction and probabilistic soft clustering in a shared latent space. Conv-VaDE enables generative decoding of cluster prototypes into verifiable scalp topographies, replacing opaque hard partitioning with probabilistic soft assignment. A polarity invariance scheme and a four-dimensional grid search over cluster count ($K \in [3,\,20]$), latent dimensionality, network depth, and channel width are conducted to systematically reveal how each architectural design choice shapes the quality, stability, and interpretability of learned EEG microstate representations. The model is evaluated on the LEMON resting-state eyes-closed EEG dataset with ten participants using topographic template formation, clustering stability, and global explained variance (GEV). The architecture search reveals that depth $L\!=\!4$ appears consistently across all 18 best-performing configurations, yielding a best-case GEV of 0.730 and a silhouette of 0.229 at $K\!=\!4$ across the model sweeps, where moderately deep networks with compact channel widths and small latent dimensionality dominate across the full $K$ range. These results establish that principled architecture search, rather than model scale, is the key to interpretable and stable EEG microstate discovery via variational deep embedding.
\end{abstract}

\keywords{EEG microstates \and Variational autoencoder \and Gaussian mixture model \and Polarity invariance \and Hyperparameter sweep \and Clustering validation \and Mechanistic Interpretability}

\section{Introduction} \label{sec:intro}

EEG microstates are brief, quasi-stable spatial distributions of scalp electric potential, typically lasting 60--120\,ms, that partition continuous brain activity into a sequence of discrete functional states~\cite{MICHEL2018577,Lehmann:2009}. Lehmann and colleagues established that microstate temporal statistics, including mean duration, occurrence rate, transition probabilities, and fractional coverage, are sensitive biomarkers, with subsequent work linking them to cognitive processing and neuropsychiatric conditions such as schizophrenia and depression~\cite{brechet2022eeg,murphy2020abnormalities, Lehmann:2009}.

Modified K-Means (ModKMeans), the predominant clustering algorithm for microstate analysis, assigns each time point to the nearest microstate class based on spatial correlation, treating opposite-sign maps as equivalent. This hard assignment discards information about transitional or ambiguous regions where the scalp topography may genuinely reflect a mixture of two or more underlying microstates. Furthermore, the convention of fixing $K\!=\!4$ microstates, while grounded in early group-level findings, may not capture the full complexity of individual brain dynamics; the optimal cluster count is known to vary across datasets, recording conditions, and clinical populations~\cite{MICHEL2018577}. Critically, ModKMeans operates directly in electrode space with no learned representation and no decoder, offering no mechanism to reconstruct or verify the cluster centres it discovers as scalp topographies. Recent deep learning architectures applied to EEG microstates~\cite{zhao2025identifying,sikka2020investigating, thukral2025convolutional} address some of these shortcomings by learning nonlinear mappings that enable reconstruction of cluster centres as scalp topographies. However, these approaches either rely on post-hoc clustering of a frozen latent space or frame microstates as supervised classification, and the challenge of determining the optimal number of microstates remains unresolved.

This gap motivates a generative approach that couples a learned decoder with a probabilistic clustering mechanism capturing assignment uncertainty. Variational Deep Embedding (VaDE)~\cite{jiang2017variational} combines a variational autoencoder with a Gaussian mixture model prior to jointly learn structured representations and probabilistic cluster assignments. The present study extends VaDE to EEG microstate analysis through a Convolutional Variational Deep Embedding (Conv-VaDE) model and investigates how cluster count, latent dimensionality, network depth, and channel width affect the quality, stability, and interpretability of learned microstate templates in a systematic four-dimensional grid search.

\section{Related Work} \label{sec:related}

\textit{Conventional microstate analysis.}
Microstate analysis was formalised by Lehmann et al.~\cite{Lehmann:2009} and systematised by Michel and Koenig~\cite{MICHEL2018577}, who established the pipeline of bandpass filtering, Global Field Power (GFP) peak extraction, spatial clustering, and temporal label assignment that underpins contemporary microstate research. Four canonical classes (usually referred to as A--D) were identified through large-scale normative studies~\cite{koenig1999deviant, KOENIG200241} and remain the standard reference. Modified K-Means and Atomize and Agglomerate Hierarchical Clustering are the predominant clustering algorithms~\cite{Poulsen289850}, both operating in electrode space with hard assignments, whereby each time point is assigned to exactly one microstate. These methods achieve polarity invariance through their distance metric, typically the absolute spatial correlation, rather than through the representation itself; a scalp topography and its sign-inverted counterpart are treated as representing the same microstate during assignment, meaning the algorithm handles polarity implicitly without explicitly encoding this invariance into the model. Clustering quality is conventionally assessed via Global Explained Variance (GEV), which measures how well the assigned microstate templates account for the observed scalp topography~\cite{MICHEL2018577}. The standard analysis pipeline applies Modified K-Means (ModKMeans), which fits prototype scalp voltage distributions by clustering multichannel EEG voltage vectors sampled at GFP peaks, where spatial configurations are most stable. The resulting microstate templates are then projected back onto the full continuous EEG recording, assigning every time point to its most correlated microstate map, thereby capturing the temporal extent of each microstate beyond the isolated GFP peak. While four canonical microstate classes are widely reported in the literature, this number reflects an empirical convention rather than a constraint of the algorithm itself~\cite{Poulsen289850,vonwegner2018eeg}.

\textit{Deep learning for EEG microstates.}
Deep learning architectures have been explored across multiple aspects of EEG microstate research. Thukral et al.~\cite{thukral2025convolutional} trained a convolutional autoencoder on EEG topographic maps and clustered the latent space with ModKMeans, improving silhouette and Davies-Bouldin scores over direct electrode-space clustering. Zhao et al.~\cite{zhao2025identifying} framed microstate identification as supervised sequence-to-sequence classification, achieving 74.26\% accuracy. Sikka et al.~\cite{sikka2020investigating} developed an LSTM-based autoencoder to capture complex temporal dependencies in EEG data, using the learned representations to model microstate transitions across the continuous recording. These approaches rely on post-hoc clustering procedures applied outside the network, and without a generative decoder, the internal representations cannot be reconstructed, leaving the discovered microstate templates as a black box. Additionally, none of them enforces polarity invariance during training. The absence of polarity invariance risks creating redundant clusters that split a single brain state along an irrelevant sign axis.

\textit{Deep clustering and variational methods.}
Deep clustering methods that jointly optimise representation learning and clustering within a single network offer a more coherent alternative to post-hoc clustering, where representation learning and clustering are treated as separate tasks. Deep Embedded Clustering (DEC)~\cite{xie2016unsupervised} jointly optimises representation learning and clustering but lacks a decoder, meaning it only optimises the clustering objective, leaving no room to visualise or inspect the scalp topographies. Its improved variant, Improved Deep Embedded Clustering (IDEC)~\cite{guo2017improved}, addresses this by adding a reconstruction loss and a decoder, but the decoder is deterministic rather than generative, preventing probabilistic assignment of EEG segments to microstate templates. Variational Deep Embedding (VaDE)~\cite{jiang2017variational} addresses this by embedding a Gaussian Mixture Model (GMM) prior, a probabilistic model that assumes data is generated from a mixture of $K$ Gaussian distributions, into the VAE framework, optimising a mixture-augmented evidence lower bound that trains encoder, decoder, and GMM parameters end-to-end. This is distinct from two-stage pipelines that cluster a frozen latent space post hoc, because the clustering objective shapes latent geometry during learning. Simple, Scalable, and Stable Variational Deep Clustering (S3VDC) extended VaDE with interleaved Kullback-Leibler divergence (KL) regularisation during pretraining to prevent posterior collapse, a failure mode where the approximate posterior ignores the latent code and the decoder generates from the prior alone~\cite{fu2019cyclical, cao2020simple}. Fu et al.~\cite{fu2019cyclical} showed that cyclical annealing of the KL weight mitigates this collapse by allowing the encoder to learn informative representations before the full KL penalty is applied.

\textit{Explainability perspective.}
From an explainability perspective, the distinction between intrinsic and post-hoc methods is central to this work~\cite{rudin2019stop}. Post-hoc approaches such as LIME~\cite{ribeiro2016should}, which explains individual predictions by fitting a simple interpretable model locally, and SHAP~\cite{lundberg2017unified}, which assigns each input feature a contribution value based on cooperative game theory, explain an already-trained model by approximating its decision boundary locally, but require a separate attribution step and do not guarantee faithfulness to the model's internal reasoning. Intrinsic explainability, by contrast, is built into the model architecture: the explanation is the model's own output rather than an external approximation. In the context of generative models, VAE decoders provide intrinsic explainability by mapping latent codes to data space~\cite{9166477}. Systematic architecture evaluation complements this by revealing how design choices affect model behaviour, contributing to model transparency in the sense of Lipton's taxonomy~\cite{lipton2018mythos}.

\section{Methodology} \label{sec:method}

This study adopts a secondary research approach, utilising the existing LEMON resting-state EEG dataset to construct and empirically evaluate a novel Conv-VAE-GMM architecture.\footnote{Code and sweep results are available at \url{https://github.com/fayisode/microstate-architecture-search}.} The hypothesis is: if a VaDE-based Conv-VaDE model with a GMM prior is trained on resting-state EEG data and subjected to a systematic four-dimensional architecture search over cluster count, latent dimensionality, network depth, and channel width, then moderately deep architectures with compact channel widths will consistently dominate the search space, yielding the most stable and interpretable microstate templates as evidenced by higher silhouette scores, lower Davies-Bouldin indices, higher Calinski-Harabasz indices, and higher GEV, while the generative decoder will enable direct reconstruction of each GMM component centre into a verifiable scalp topography.

\begin{figure}[!t]
  \centering
  \includegraphics[width=\textwidth]{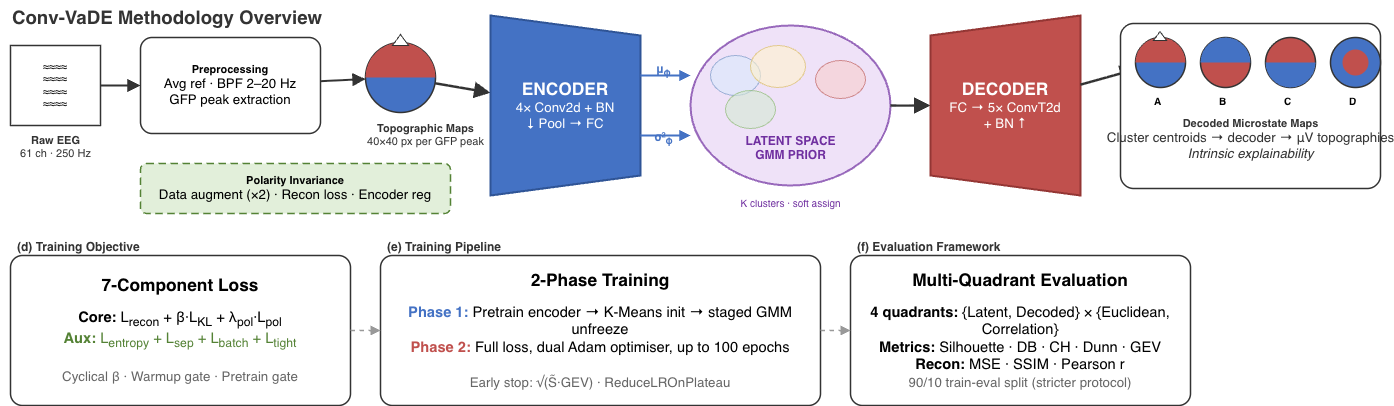}
  \caption{Conv-VaDE processes EEG recordings into topographic maps and clusters them using a convolutional encoder with a GMM-structured latent space, trained via a seven-component objective and evaluated across multiple representation and distance metrics.}
  \label{fig:methodology}
\end{figure}

\textit{Data Understanding and Preparation.}
The LEMON dataset~\cite{babayan2019mind} comprises eyes-closed resting-state EEG at $f_s\!=\!250$\,Hz, 61 channels, 10--10 montage. Ten participants were selected from the dataset. A common average reference and zero-phase FIR bandpass filter (2--20\,Hz) preserves temporal alignment and prevents phase-induced microstate boundary blurring. GFP peaks are extracted via Pycrostates~\cite{ferat2022pycrostates} (min.\ inter-peak distance: 3 samples) and projected onto $40\!\times\!40$ topographic images via azimuthal equidistant projection and cubic interpolation~\cite{ahmed2022examining}. Z-score normalisation with $\pm5\sigma$ clipping is computed on the training split only: $x_{\mathrm{norm}} = \mathrm{clip}((x - \bar{x}_{\mathrm{train}})/\sigma_{\mathrm{train}},\,-5,\,+5)$. A 90/10 train/evaluation split maximises training data; the evaluation set is left unaugmented.

\textit{Model Architecture.}
The encoder maps $\mathbf{x}\!\in\!\mathbb{R}^{1\times40\times40}$ through four \texttt{Conv2d} layers ($\texttt{ndf}\!\to\!2\texttt{ndf}\!\to\!4\texttt{ndf}\!\to\!8\texttt{ndf}$) with batch normalisation, LeakyReLU ($\alpha\!=\!0.2$), and dropout ($p\!=\!0.2$), then adaptive average pooling and FC layers producing $(\boldsymbol{\mu}_\phi,\log\boldsymbol{\sigma}^2_\phi)\!\in\!\mathbb{R}^{d_z}$ via reparameterisation~\cite{kingma2022autoencodingvariationalbayes}. The decoder mirrors this with \texttt{ConvTranspose2d} and a linear final activation. The GMM prior~\cite{jiang2017variational} $p(\mathbf{z})\!=\!\sum_{k=1}^{K}\pi_k\mathcal{N}(\mathbf{z};\boldsymbol{\mu}_k^{(c)},\mathrm{diag}(\boldsymbol{\sigma}_k^{2(c)}))$ is optimised end-to-end. The search varies $K\!\in\!\{3,\ldots,20\}$, $d_z\!\in\!\{16,32,64\}$, depth$\,\in\!\{2,3,4\}$, $\texttt{ndf}\!\in\!\{32,64,128\}$, yielding 486 unique configurations evaluated across 10 subjects.

\textit{Polarity Invariance and Training Objective.}
Polarity invariance ($\mathbf{x}$ and $-\mathbf{x}$ are the same state) is enforced at three levels: (1)~sign-flip augmentation of the training set, with GMM initialisation on the unaugmented subset; (2)~polarity-invariant loss $\mathcal{L}_{\mathrm{recon}}\!=\!\min(\mathrm{MSE}(\hat{\mathbf{x}},\mathbf{x}),\mathrm{MSE}(\hat{\mathbf{x}},-\mathbf{x}))\cdot D_x$, $D_x\!=\!1600$; (3)~encoder regularisation $\mathcal{L}_{\mathrm{pol}}\!=\!\mathrm{MSE}(\boldsymbol{\mu}_\phi(\mathbf{x}),\boldsymbol{\mu}_\phi(-\mathbf{x}))$, $\lambda_{\mathrm{pol}}\!=\!0.1$.
The total loss is $\mathcal{L} = \mathcal{L}_{\mathrm{recon}} + \beta\mathcal{L}_{\mathrm{KL}} + \lambda_e\mathcal{L}_{\mathrm{entropy}} + \lambda_s\mathcal{L}_{\mathrm{sep}} + \lambda_b\mathcal{L}_{\mathrm{batch}} + \lambda_t\mathcal{L}_{\mathrm{tight}} + \lambda_{\mathrm{pol}}\mathcal{L}_{\mathrm{pol}}$, where $\mathcal{L}_{\mathrm{entropy}}$ ($\lambda_e\!=\!0.3/\log K$) prevents weight collapse; $\mathcal{L}_{\mathrm{sep}}$ ($\lambda_s\!=\!50.0$) penalises topographic redundancy via mean pairwise $|\mathrm{Pearson}\,r|$; $\mathcal{L}_{\mathrm{batch}}$ ($\lambda_b\!=\!5.0$) enforces uniform cluster usage; $\mathcal{L}_{\mathrm{tight}}$ ($\lambda_t\!=\!0.2$) pulls samples toward assigned cluster priors~\cite{cao2020simple}. Cyclical $\beta$-annealing ($\beta_{\max}\!=\!0.1$, 4~cycles)~\cite{fu2019cyclical} modulates KL pressure; auxiliary losses are zeroed for 3~epochs then linearly ramped over 10.

\textit{Training Procedure.}
Pretraining runs 200 steps ($\beta\!=\!10^{-3}$, no auxiliary losses), followed by GMM initialisation via bisecting K-Means on unaugmented latent means (flat for $K\!\leq\!4$, iterative bisection for $K\!>\!4$), prior frozen for 5~epochs, dead clusters ($<\!1\%$) reinitialised from the largest surviving cluster. Main training runs up to 100~epochs with dual Adam optimisers (encoder/decoder: $\eta\!=\!10^{-3}$, $\lambda_{L2}\!=\!10^{-5}$; GMM: $\eta\!=\!5\!\times\!10^{-4}$), gradient clipping at 5.0 and 1.0, and early stopping on $\sqrt{\tilde{S}\cdot\mathrm{GEV}}$ ($\tilde{S}\!=\!(\mathrm{sil}+1)/2$), with ReduceLROnPlateau (patience\,$=\!10$).

\textit{Evaluation.}
Clustering quality is assessed in the latent space via Silhouette, Davies--Bouldin, Calinski--Harabasz, and Dunn indices. Reconstruction quality is measured with MSE, SSIM (range\,$=\!10.0$), and spatial correlation. Topographic fidelity is evaluated via GEV computed through backfitting: $\mathrm{GEV}\!=\!\sum_t\mathrm{GFP}^2(t)\rho^2(t)/\sum_t\mathrm{GFP}^2(t)$. For each $K$, the best configuration is selected by the highest mean GEV averaged across all 10 subjects.

\section{Results and Discussion} \label{sec:results}

\begin{figure}[!t]
  \centering
  \begin{subfigure}{0.24\textwidth}
    \includegraphics[width=\linewidth]{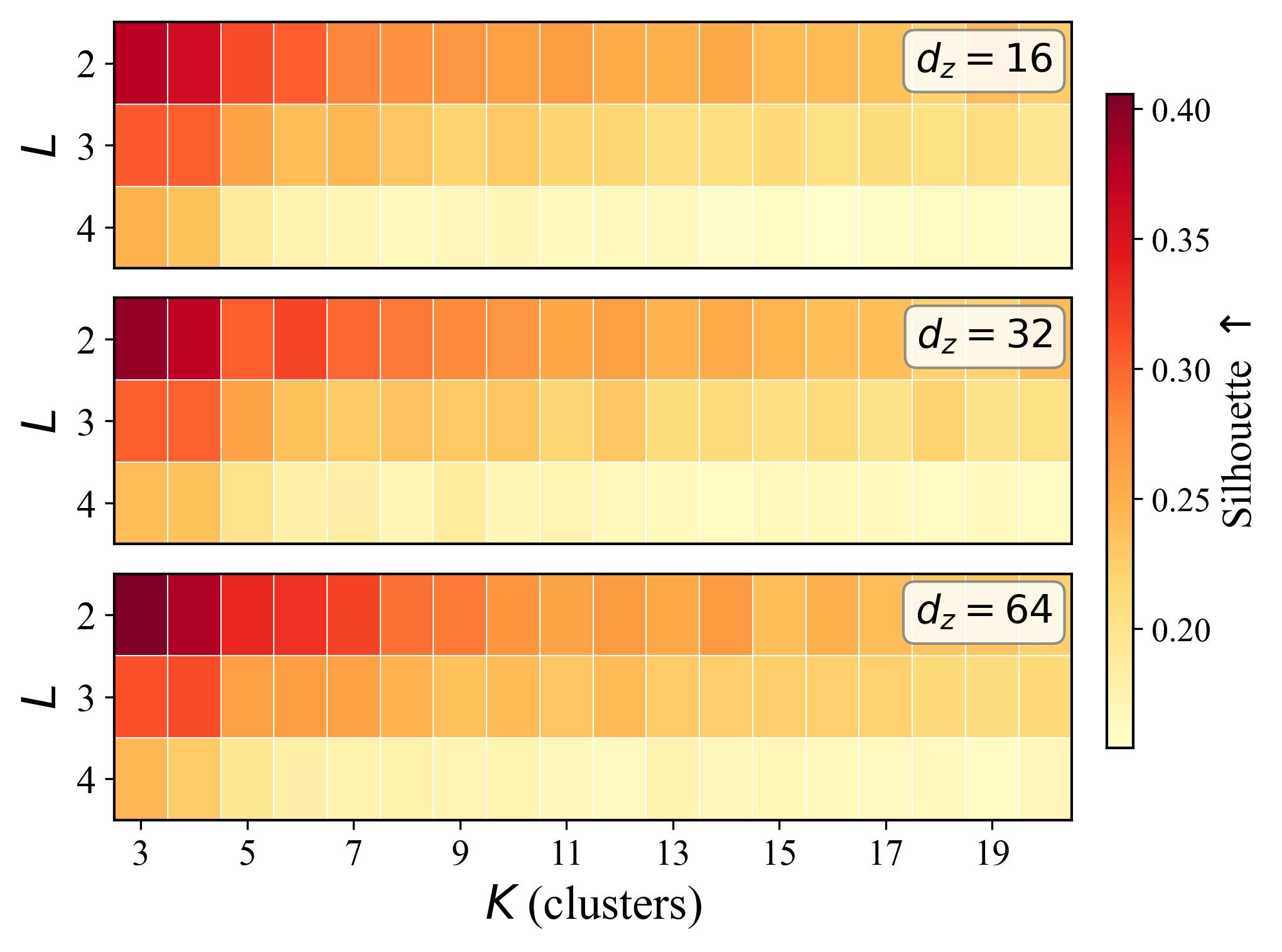}
    \caption{Silhouette}
  \end{subfigure}\hfill
  \begin{subfigure}{0.24\textwidth}
    \includegraphics[width=\linewidth]{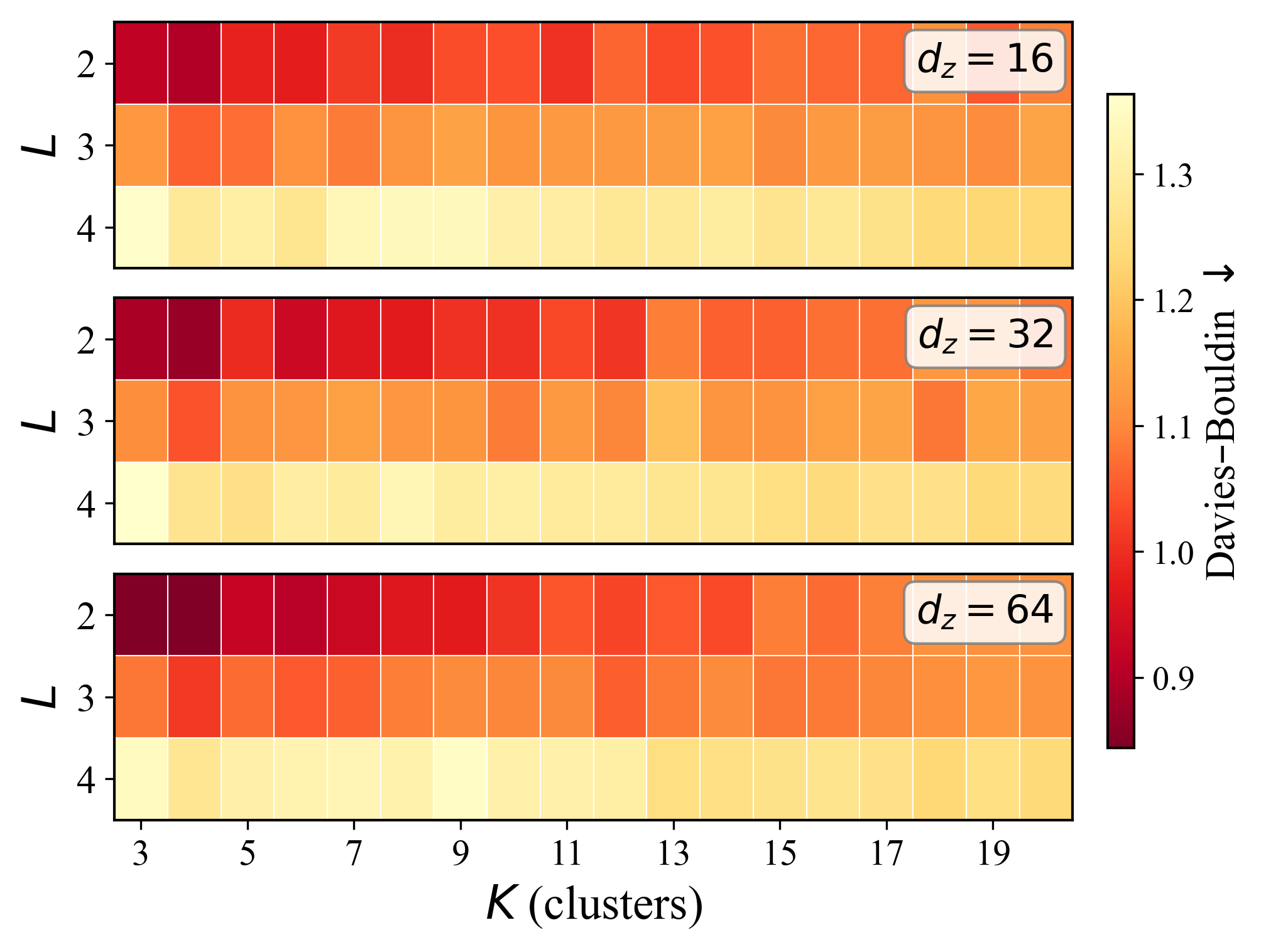}
    \caption{Davies--Bouldin}
  \end{subfigure}\hfill
  \begin{subfigure}{0.24\textwidth}
    \includegraphics[width=\linewidth]{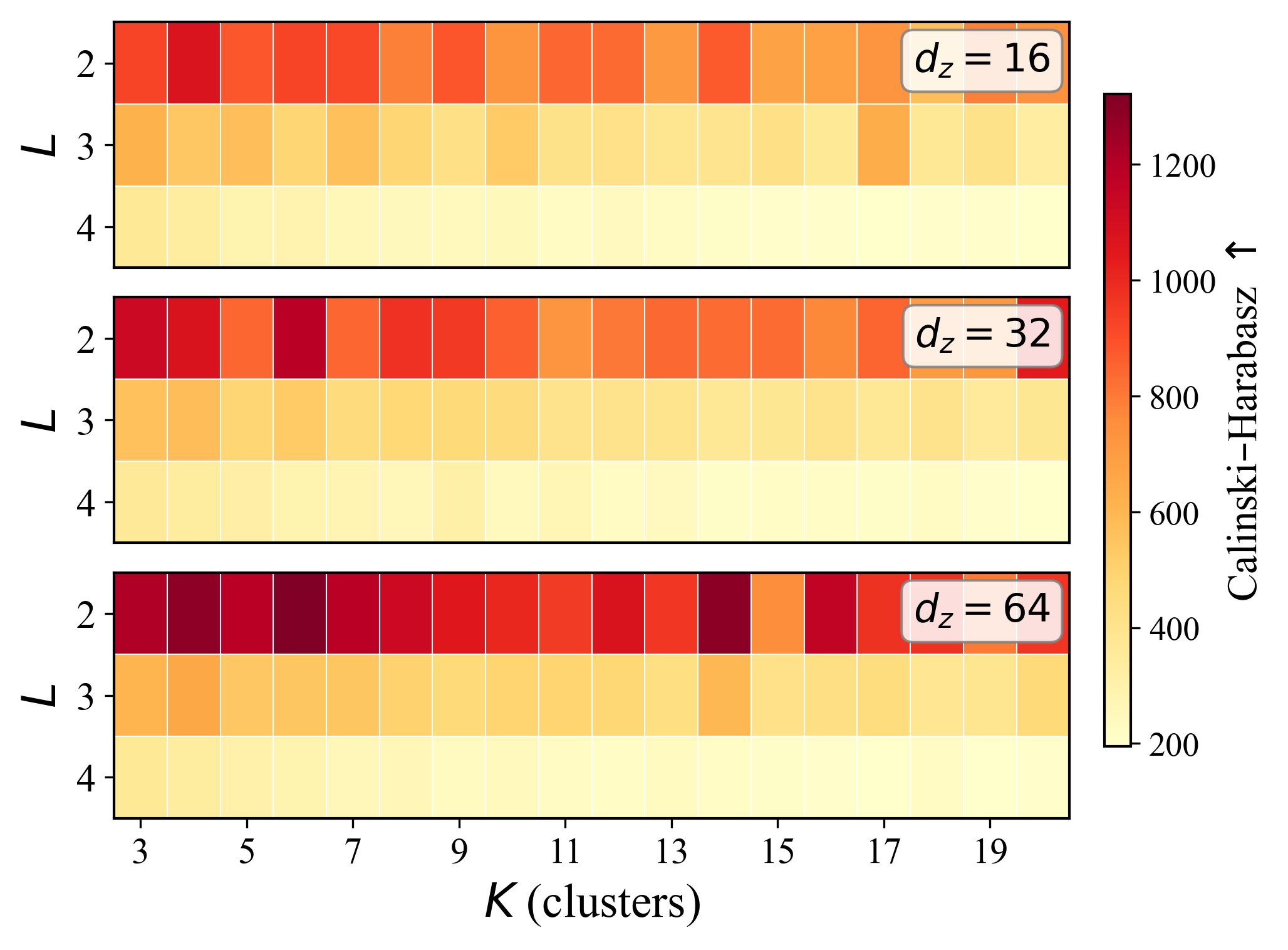}
    \caption{Calinski--Harabasz}
  \end{subfigure}\hfill
  \begin{subfigure}{0.24\textwidth}
    \includegraphics[width=\linewidth]{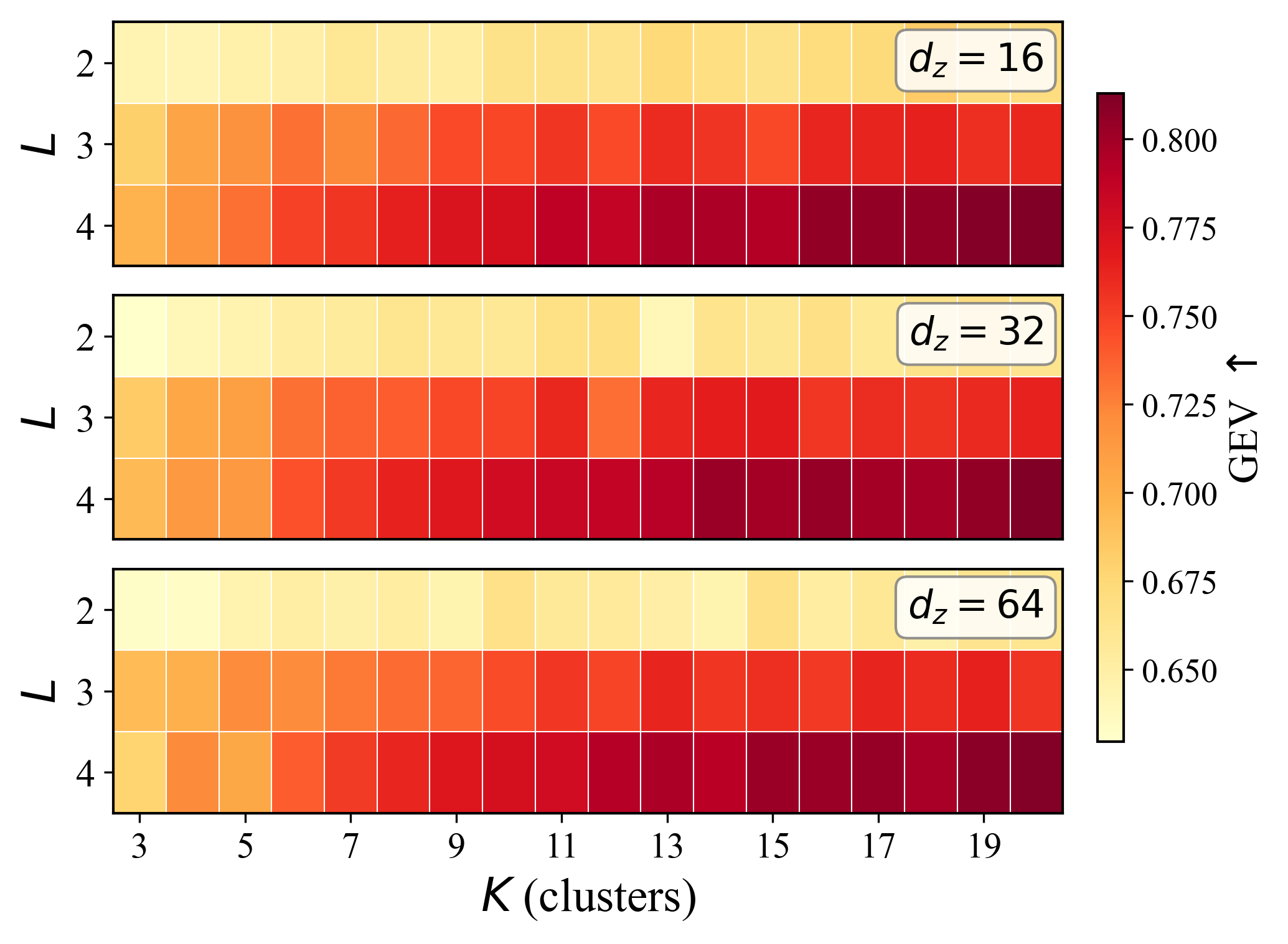}
    \caption{GEV}
  \end{subfigure}
  \caption{Q1 (latent Euclidean) clustering metric landscapes across the $K \times L \times d_z$ sweep space. Each heatmap panel is faceted by $d_z$, with rows showing depth $L$ and columns showing cluster count $K$; $n_f$ is averaged. Colour intensity encodes metric magnitude.}
  \label{fig:cube_clean}
\end{figure}

\begin{figure}[!t]
  \centering
  \begin{subfigure}{0.48\textwidth}
    \includegraphics[width=\linewidth]{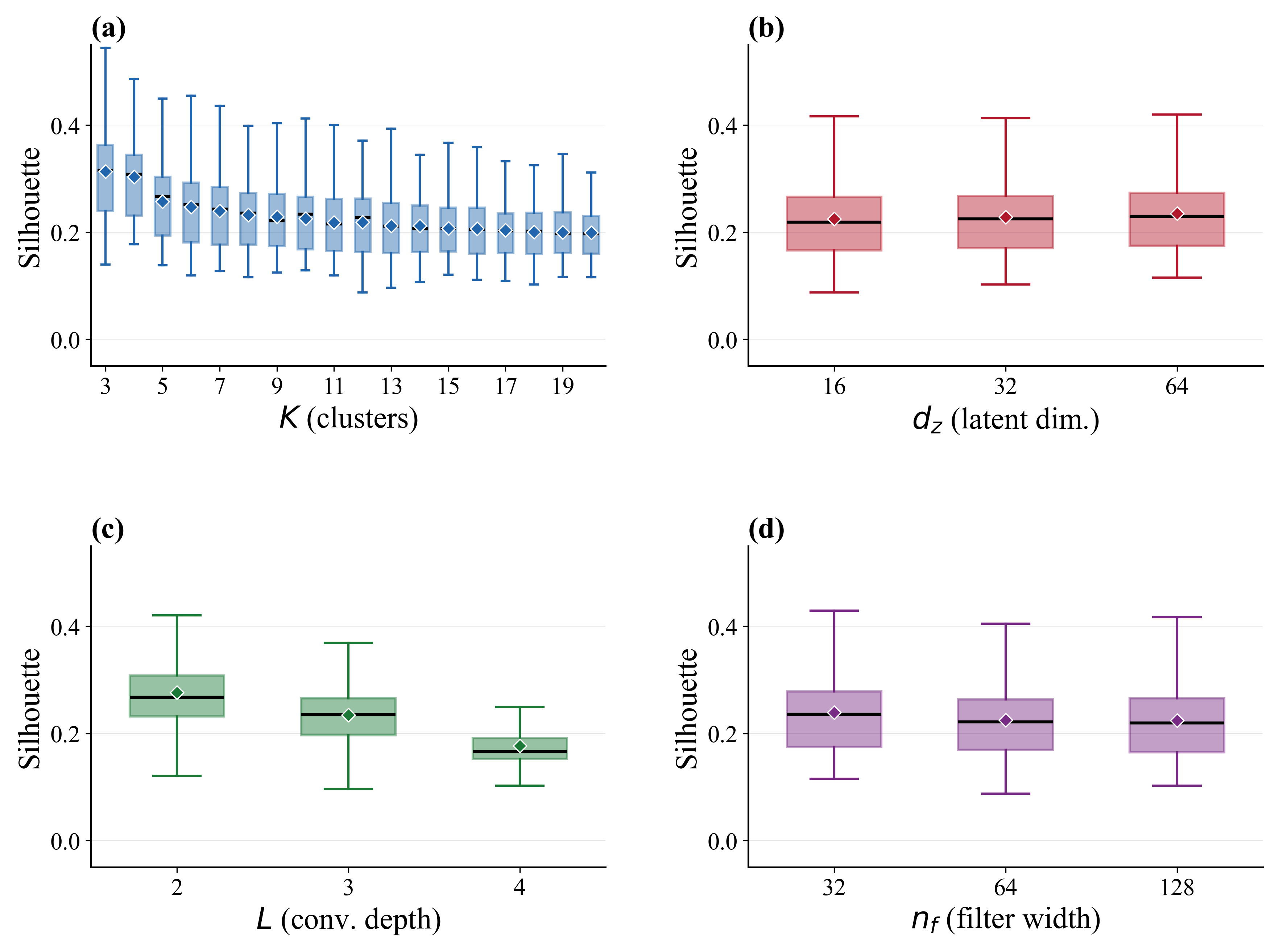}
    \caption{Silhouette}
  \end{subfigure}\hfill
  \begin{subfigure}{0.48\textwidth}
    \includegraphics[width=\linewidth]{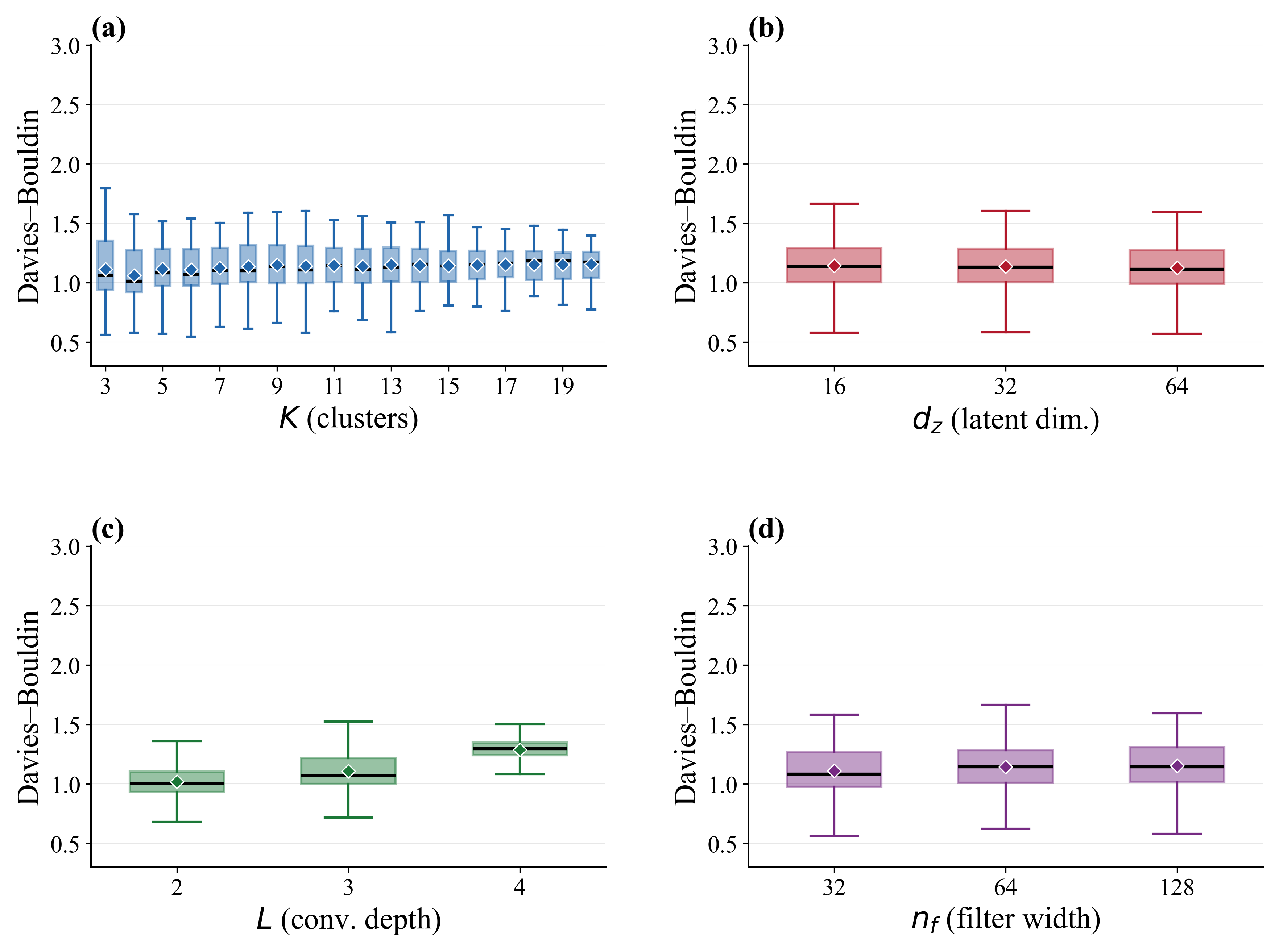}
    \caption{Davies--Bouldin}
  \end{subfigure}\\[1ex]
  \begin{subfigure}{0.48\textwidth}
    \includegraphics[width=\linewidth]{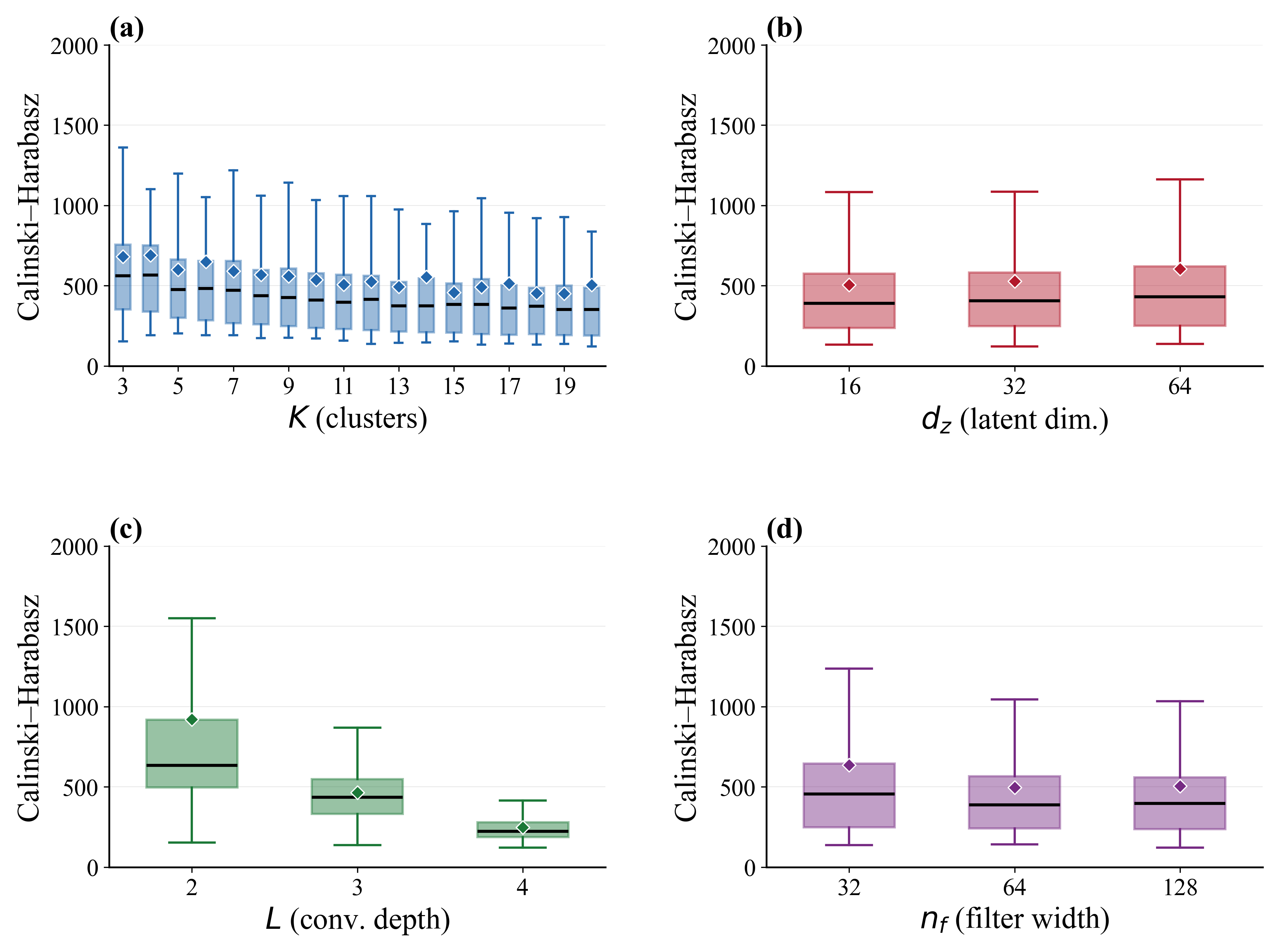}
    \caption{Calinski--Harabasz}
  \end{subfigure}\hfill
  \begin{subfigure}{0.48\textwidth}
    \includegraphics[width=\linewidth]{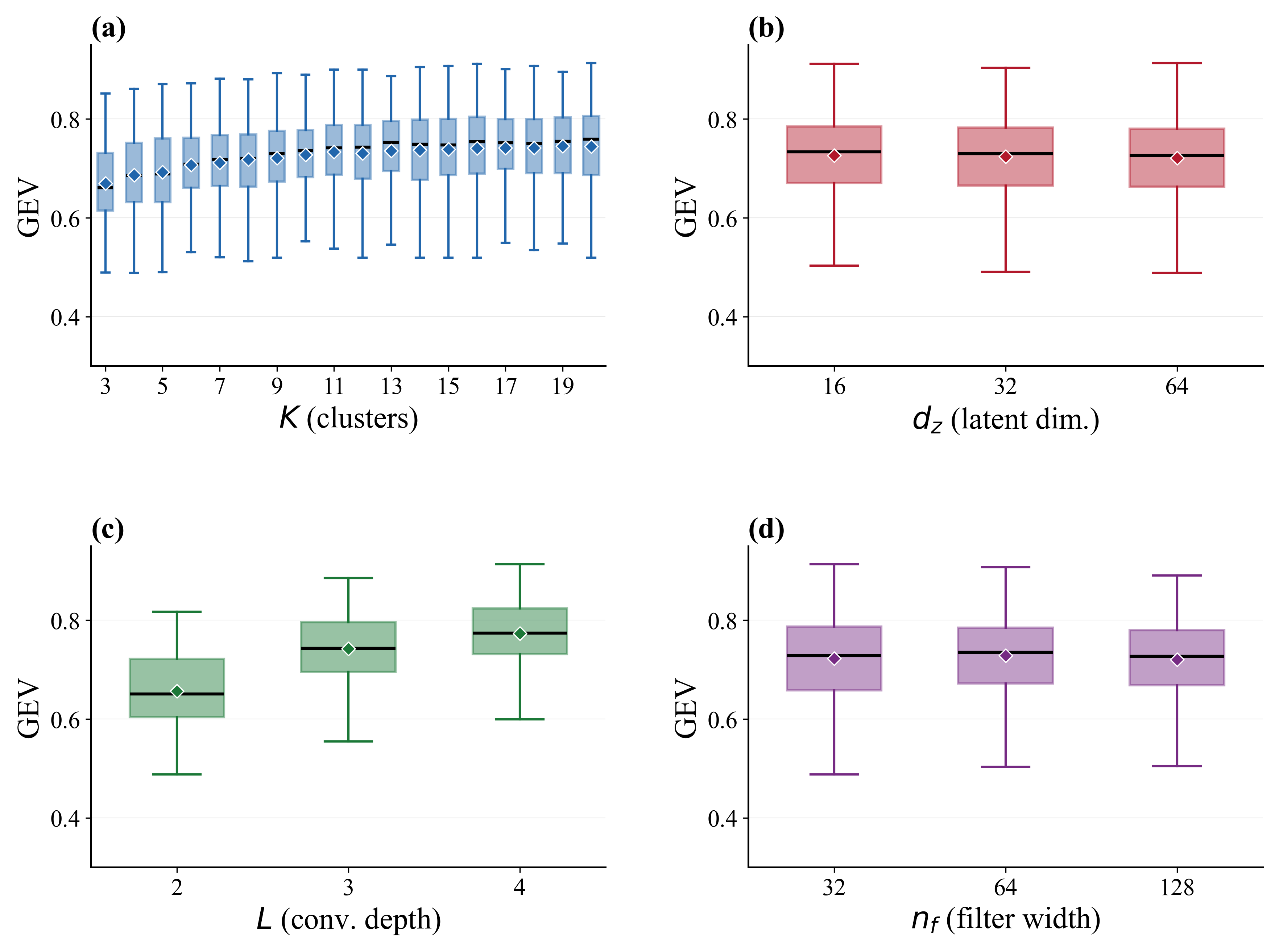}
    \caption{GEV}
  \end{subfigure}
  \caption{Architecture parameter effects on Q1 (latent Euclidean) clustering metrics. Each panel shows the marginal effect of one sweep dimension ($d_z$, $L$, $n_f$) on a given metric, aggregated across all $K$ values.}
  \label{fig:arch_effects}
\end{figure}

\begin{figure}[!t]
  \centering
  \includegraphics[width=0.85\textwidth]{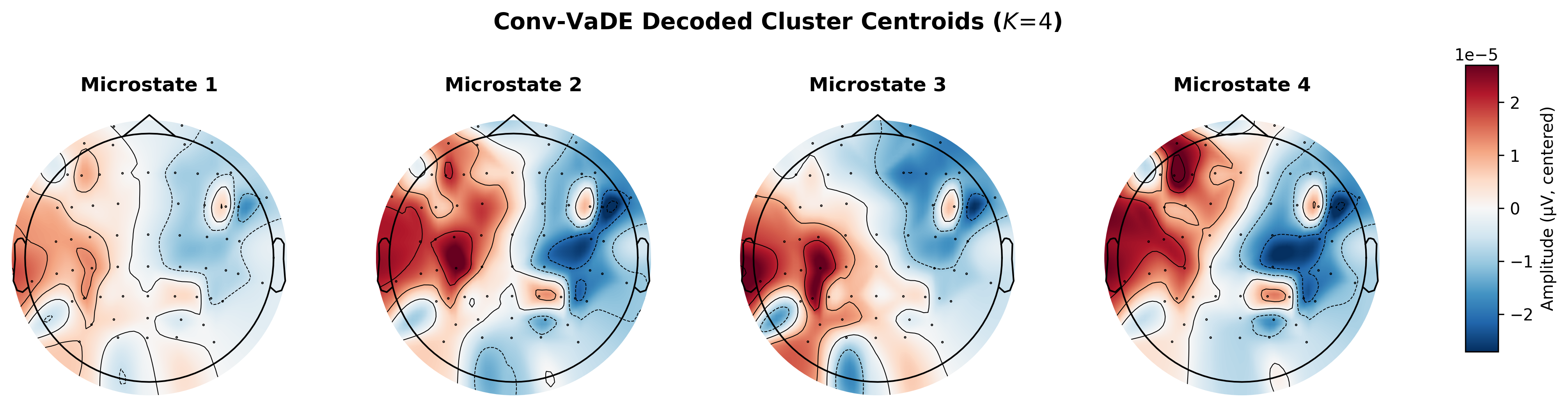}
  \caption{Decoded GMM cluster centroids at $K\!=\!4$ ($d_z\!=\!16$, $L\!=\!4$, $n_f\!=\!32$, subject 010012). Each panel shows the decoder output $\hat{\mathbf{x}}_k = g_\theta(\boldsymbol{\mu}_k^{(c)})$ for cluster $k$, rendered as a circular scalp topography. Colour encodes z-scored amplitude (RdBu\_r).}
  \label{fig:centroids}
\end{figure}

\begin{figure}[!t]
  \centering
  \begin{subfigure}{0.24\textwidth}
    \includegraphics[width=\linewidth]{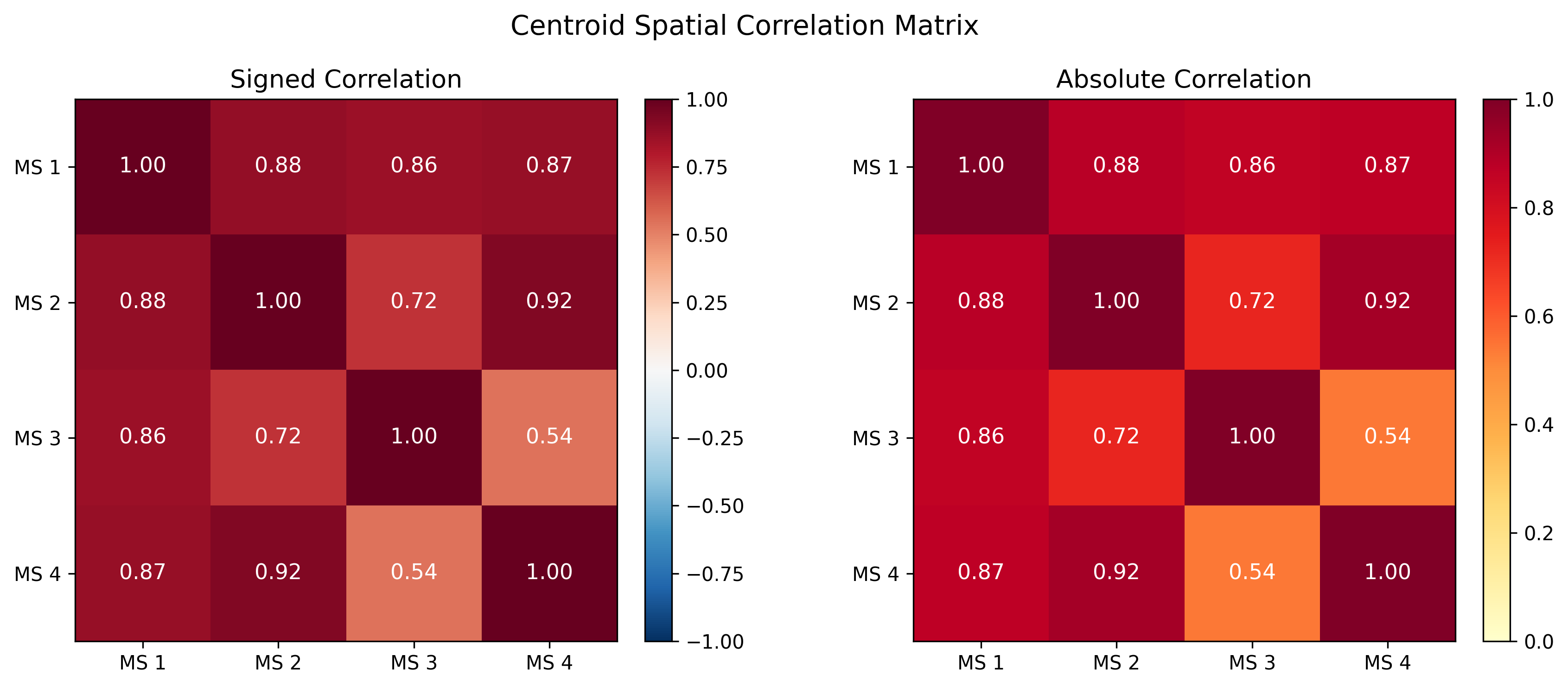}
    \caption{Centroid correlation}
    \label{fig:corr_matrix}
  \end{subfigure}\hfill
  \begin{subfigure}{0.24\textwidth}
    \includegraphics[width=\linewidth]{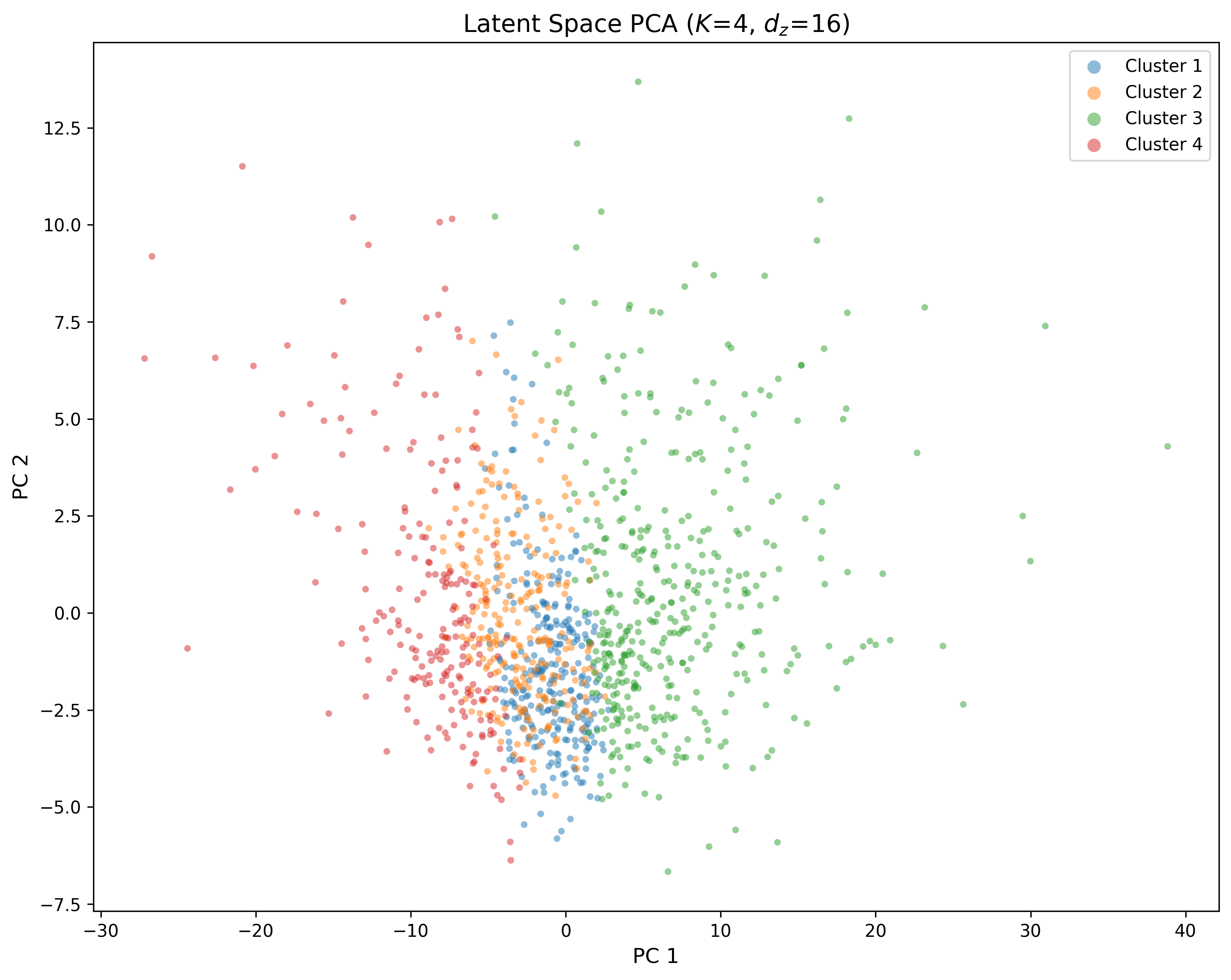}
    \caption{Latent PCA}
    \label{fig:latent_pca}
  \end{subfigure}\hfill
  \begin{subfigure}{0.24\textwidth}
    \includegraphics[width=\linewidth]{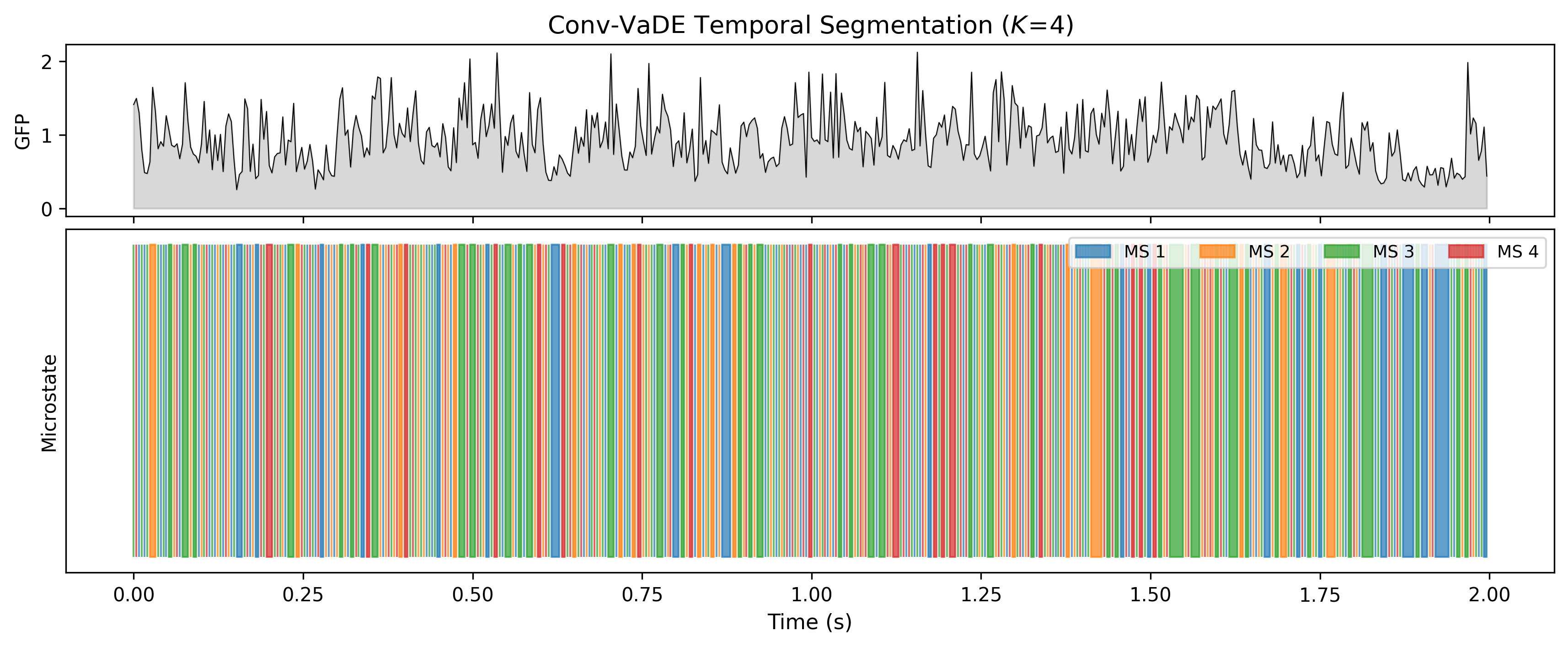}
    \caption{Segmentation}
    \label{fig:segmentation}
  \end{subfigure}\hfill
  \begin{subfigure}{0.24\textwidth}
    \includegraphics[width=\linewidth]{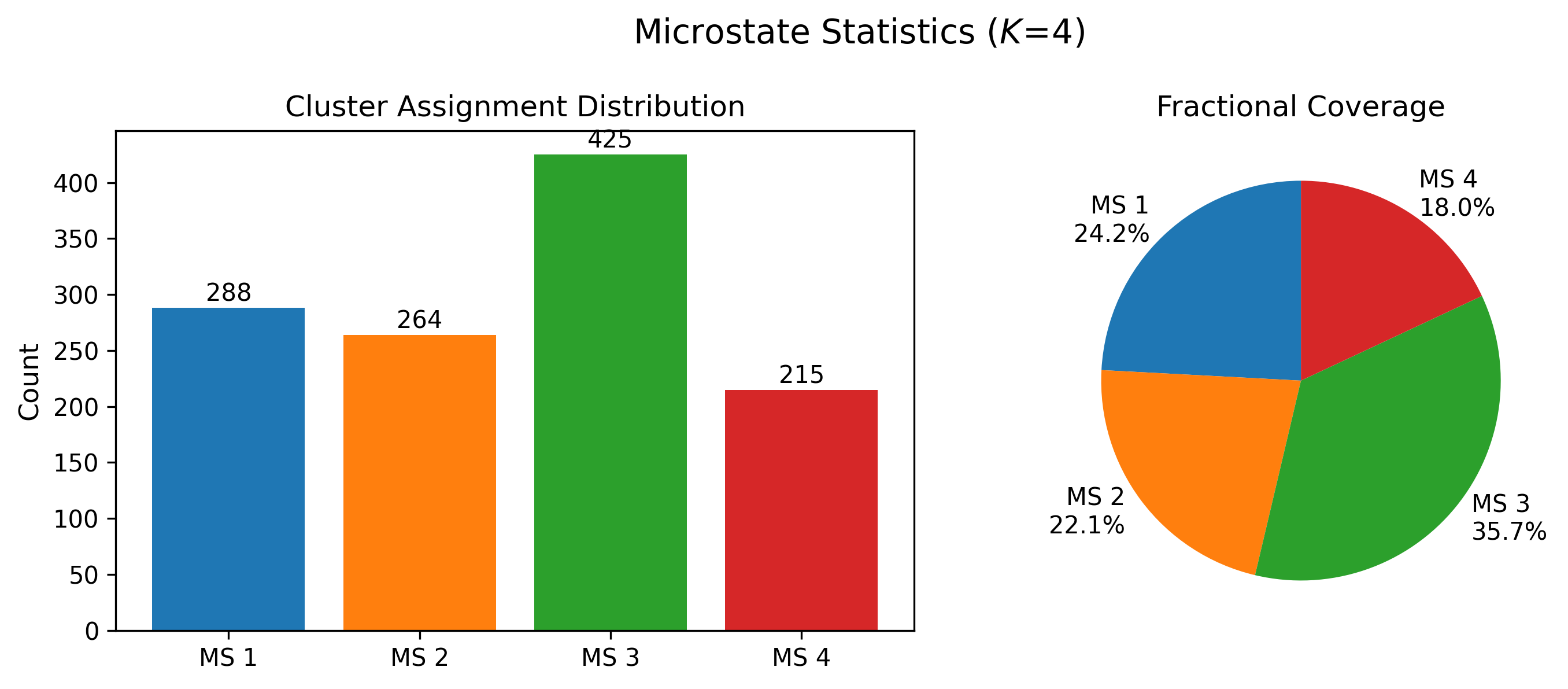}
    \caption{Cluster distribution}
    \label{fig:stats}
  \end{subfigure}
  \caption{Explainability analysis at $K\!=\!4$ (subject 010012). (a)~Signed Pearson correlation between decoded centroids. (b)~PCA of the 16D latent space coloured by GMM assignment. (c)~GFP trace and microstate ribbon showing temporal clustering. (d)~Cluster sample counts and fractional coverage.}
  \label{fig:xai_panel}
\end{figure}

Table~\ref{tab:sweep} reports the best configuration per $K$ from the architecture sweep, selected by highest mean GEV across 10 subjects. Figures~\ref{fig:cube_clean} and~\ref{fig:arch_effects} visualise Q1 clustering metric landscapes and architecture parameter effects across the sweep space. Figures~\ref{fig:centroids} and~\ref{fig:xai_panel} demonstrate the explainability of Conv-VaDE through decoded cluster centroids, latent space structure, temporal clustering, and cluster distribution analysis. At $K\!=\!4$, the best architecture ($d_z\!=\!16$, $L\!=\!4$, $n_f\!=\!32$) achieves a silhouette of $0.229 \pm 0.039$ and a Davies--Bouldin index of $1.28 \pm 0.13$, indicating well-separated and compact clusters in the learned latent space, with a global explained variance of $0.730 \pm 0.072$ confirming strong topographic representational fidelity in electrode space.

\begin{table}[!t]
    \caption{Architecture sweep: best configuration per $K$ by GEV (held-out 10\% split, $n\!=\!10$ subjects). Q1 Sil = latent Euclidean silhouette.}
    \label{tab:sweep}
    \centering
    \small
    \setlength{\tabcolsep}{4pt}
    \begin{minipage}{0.48\textwidth}
        \centering
        \begin{tabular}{cccccc}
        \toprule
        $K$ & $d_z$ & Depth & \texttt{ndf} & Q1 Sil $\uparrow$ & GEV $\uparrow$ \\
        \midrule
        3  & 16 & 4 & 32  & \textit{0.234} & 0.710 \\
        4  & 16 & 4 & 32  & 0.229 & 0.730 \\
        5  & 16 & 4 & 32  & 0.195 & 0.741 \\
        6  & 16 & 4 & 32  & 0.165 & 0.760 \\
        7  & 16 & 4 & 32  & 0.171 & 0.771 \\
        8  & 16 & 4 & 32  & 0.167 & 0.777 \\
        9  & 32 & 4 & 32  & 0.186 & 0.786 \\
        10 & 16 & 4 & 32  & 0.162 & 0.792 \\
        11 & 16 & 4 & 32  & 0.158 & 0.798 \\
        \bottomrule
        \end{tabular}
    \end{minipage}\hfill
    \begin{minipage}{0.48\textwidth}
        \centering
        \begin{tabular}{cccccc}
        \toprule
        $K$ & $d_z$ & Depth & \texttt{ndf} & Q1 Sil $\uparrow$ & GEV $\uparrow$ \\
        \midrule
        12 & 64 & 4 & 32  & 0.162 & 0.803 \\
        13 & 16 & 4 & 32  & 0.172 & 0.806 \\
        14 & 32 & 4 & 32  & 0.160 & 0.814 \\
        15 & 16 & 4 & 32  & 0.161 & 0.810 \\
        16 & 16 & 4 & 32  & 0.150 & 0.817 \\
        17 & 64 & 4 & 32  & 0.164 & 0.813 \\
        18 & 16 & 4 & 32  & 0.161 & 0.817 \\
        19 & 16 & 4 & 32  & 0.165 & 0.817 \\
        20 & 16 & 4 & 32  & 0.157 & \textit{0.821} \\
        \bottomrule
        \end{tabular}
    \end{minipage}
\end{table}

\textit{Explainability of Decoded Centroids.}
The core explainability claim of Conv-VaDE is that GMM cluster centres can be decoded into inspectable scalp topographies. Figure~\ref{fig:centroids} shows the four decoded centroids at $K\!=\!4$, each rendered as a circular scalp map in the original $40\!\times\!40$ pixel grid. The topographies reveal distinct spatial patterns consistent with known microstate classes: anterior-posterior gradients and lateral asymmetries are clearly visible, confirming that the decoder has learned a meaningful mapping from latent cluster centres to physiologically plausible scalp distributions.

Figure~\ref{fig:xai_panel} provides further explainability evidence. The centroid correlation matrix~(\subref{fig:corr_matrix}) shows moderate-to-high absolute correlations ($|r|\!=\!0.54$--$0.92$), indicating shared global topographic structure while signed differences capture polarity contrasts. The PCA projection~(\subref{fig:latent_pca}) confirms that the GMM prior imposes separable cluster structure in the first two principal components. The temporal clustering~(\subref{fig:segmentation}) reveals characteristic rapid alternation between states consistent with the 60--120\,ms durations reported in the literature~\cite{MICHEL2018577}. The approximately balanced cluster distribution~(\subref{fig:stats}) confirms that batch entropy regularisation prevents mode collapse.

\textit{Discussion.}
The multi-quadrant framework reveals an asymmetry a single metric would obscure. ModKMeans templates are native electrode vectors optimised for spatial correlation, giving a structural advantage on GEV. Conv-VaDE decoded centroids must transit from the $40\!\times\!40$ pixel grid back to 61 electrodes before GEV is computed, attenuating the correlation that GEV weights quadratically. Comparing Q4 places both methods on equal footing in decoded topographic space; Q1 and Q2 characterise latent quality that ModKMeans cannot provide.

The polarity scheme addresses a failure visible at high $K$: without it, the GMM allocates separate components to each polarity, halving the useful cluster count. Q1 silhouette generally declines from 0.234 at $K\!=\!3$ to 0.157 at $K\!=\!20$ (Table~\ref{tab:sweep}), though the decline is not strictly monotonic, indicating that latent cluster separation degrades as the number of states increases. The polarity scheme eliminates mirror-image duplicates but does not prevent semantic splitting of related brain states.

The architecture sweep fills a gap absent from prior work: optimal $d_z$, depth, and channel width for topographic EEG data cannot be assumed from image benchmarks. Depth $L\!=\!4$ dominates all 18 top-ranked configurations across every $K$ value. Latent dimensionality shows low sensitivity: $d_z\!=\!16$ dominates in 14 of 18 best configurations across the full $K$ range, with $d_z\!\in\!\{32,64\}$ appearing only at $K\!\in\!\{9,12,14,17\}$, suggesting that the $40\!\times\!40$ topographic input is sufficiently represented by a compact latent space regardless of cluster count. Channel width $n_f\!=\!32$ suffices in all 18 best configurations, confirming that the $40\!\times\!40$ topographic input does not benefit from wider convolutional filters.

\section{Conclusion} \label{sec:conclusion}

This paper presented Conv-VaDE, a convolutional variational deep embedding model for explainable EEG microstate analysis that jointly learns topographic reconstruction and probabilistic soft clustering. Decoded GMM cluster centres produce inspectable scalp topographies without post-hoc processing (Figure~\ref{fig:centroids}), and the structured latent space enables direct visualisation of cluster geometry (Figure~\ref{fig:xai_panel}). Three contributions were introduced: a three-level polarity invariance scheme eliminating redundant clusters; a seven-component objective preventing mode collapse across $K\!\in\![3,\,20]$; and a four-dimensional architecture sweep over 486 configurations evaluated through a multi-quadrant metric framework on 10 subjects. Population-level generalisation requires validation on larger cohorts. Future work targets latent-space backfitting to close the GEV gap, multi-subject generalisation, and temporal transition modelling. The source code, trained model configurations, and sweep results are publicly available at \url{https://github.com/fayisode/microstate-architecture-search}.

\section*{Acknowledgments}
This research was conducted at the Artificial Intelligence and
Cognitive Load Research Lab, University College Cork, and supported
by Research Ireland - Taighde \'Eireann, under Grant Nos. 18/CRT/6223 and 12/RC/2289-P2, co-funded by the European Regional Development Fund. For the purpose of Open Access, the author has applied a CC BY public copyright licence to any Author Accepted Manuscript version arising from this submission.

\section*{Declaration on Generative AI}
The authors used Grammarly to help with grammar check, spelling correction, and stylistic clarity during the preparation of this manuscript.

\bibliographystyle{unsrtnat}
\bibliography{references}

\appendix

\end{document}